\documentclass[letterpaper, 10 pt, conference]{ieeeconf}  
\usepackage{subcaption} 
\usepackage{xcolor}
\usepackage{caption}
\usepackage{wrapfig}
\usepackage{graphics}
\usepackage[ruled,vlined]{algorithm2e}
\usepackage[noend]{algpseudocode}
\usepackage{multirow}
\usepackage{adjustbox}
\usepackage{overpic}
\usepackage[list=off]{caption}

\makeatletter\newcommand{\manuallabel}[2]{\def\@currentlabel{#2}\label{#1}}\makeatother

\usepackage{amssymb}
\usepackage{mathtools}
\usepackage{amsmath}

\mathchardef\mhyphen="2D   

\newcommand{\R}     {\mathbb{R}}          
\newcommand{\T}     {\top}                
\newcommand{\I}     {\mathbf{I}}          

\newcommand{\estimated} [1]{\tilde{#1}}

\newcommand{\pinv}      [1]{{#1}^\dagger}

\newcommand{\nd}      {n}                              
\newcommand{\Nd}      {\mathcal{\MakeUppercase{\nd}}}  

\newcommand{\bx}     {\mathbf{x}}         

\newcommand{\nx}     {p}                  
\newcommand{\dimx}   {\mathcal{\MakeUppercase{\nx}}} 

\newcommand{\btheta}  {\boldsymbol{\theta}}               
\newcommand{\ntheta}  {p}                                 
\newcommand{\dimtheta}{\mathcal{\MakeUppercase{\ntheta}}} 

\newcommand{\bLambda}     {\mathbf{\Lambda}}         
\newcommand{\bA}     {\mathbf{A}}         
\newcommand{\bb}     {\mathbf{b}} 

\newcommand{\ebb}     {\estimated{\mathbf{b}}}

\newcommand{\nb}     {s}                  
\newcommand{\dimb}   {\mathcal{\MakeUppercase{\nb}}} 
\newcommand{\bN}     {\mathbf{N}}         
\newcommand{\bw}     {\mathbf{w}}         
\newcommand{\bv}     {\mathbf{v}}         
\newcommand{\bPhi}     {\mathbf{\Phi}}         
\newcommand{\bq}     {\mathbf{q}}         
\newcommand{\br}     {\mathbf{r}}         
\newcommand{\bJ}     {\mathbf{J}}         
\newcommand{\bqdot}  {\dot{\bq}}          

\newcommand{\bB}     {\mathbf{B}}         

\newcommand  {\bu}  {\mathbf{u}}          
\renewcommand{\nu}  {q}                   
\newcommand  {\dimu}{\mathcal{\MakeUppercase{\nu}}} 
\newcommand  {\bpi} {\boldsymbol{\pi}}    

\newcommand{\bxn}    {\bx_{\nd}}          

\newcommand  {\ebA}     {\estimated{\bA}}  
\newcommand  {\ebN}     {\estimated{\bN}}  

\renewcommand{\a}       {a}                %
\newcommand  {\ba}      {\boldsymbol{\alpha}}  %
\newcommand{\eg}{\textit{e.g.,}~} %
\newcommand{\ie}{\textit{i.e.,}~} %
\newcommand{\etal}{\textit{et al.}\xspace} %

\newcommand*{\sref}[1]{\S\ref{s:#1}}            
\newcommand*{\tref}[1]{\tablename~\ref{t:#1}}   
\newcommand*{\fref}[1]{\figurename~\ref{f:#1}}  
\newcommand*{\eref}[1]{(\ref{e:#1})}            

\let\labelindent\relax 
\usepackage[inline]{enumitem}
\setlist{nolistsep}
\newcommand{\il}[1]{\begin{enumerate*}[label=(\roman*)]#1\end{enumerate*}}

\providecommand{\provideabbreviation}[2]{\providecommand{#1}{#2\xspace}} 
\newcommand{\newabbreviation}[2]{#1\ (\renewcommand{#1}{#2\xspace}#1)}

\newcommand{\PbD    }{programming by demonstration\xspace}
\newcommand{\DoF    }{degrees of freedom\xspace}

\providecommand{\NPPE}{normalised \PPE}
\providecommand{\NPOE}{normalised \POE}
\provideabbreviation{\RULA}{Rapid Upper Limb Assessment}
\provideabbreviation{\NMSE}{normalised mean squared error}
\usepackage{amssymb}
\usepackage{mathtools}

\mathchardef\mhyphen="2D   

\providecommand{\R}     {\mathbb{R}}          
\providecommand{\T}     {\top}                
\providecommand{\I}     {\mathbf{I}}          

\providecommand{\estimated} [1]{\tilde{#1}}

\providecommand{\pinv}      [1]{{#1}^\dagger}

\providecommand{\nd}      {n}                              
\providecommand{\Nd}      {\mathcal{\MakeUppercase{\nd}}}  

\providecommand{\bx}     {\mathbf{x}}         
\providecommand{\nx}     {p}                  
\providecommand{\dimx}   {\mathcal{\MakeUppercase{\nx}}} 

\providecommand{\btheta}  {\boldsymbol{\theta}}               
\providecommand{\ntheta}  {p}                                 
\providecommand{\dimtheta}{\mathcal{\MakeUppercase{\ntheta}}} 

\providecommand{\bA}     {\mathbf{A}}         
\providecommand{\bb}     {\mathbf{b}}         
\providecommand{\nb}     {s}                  
\providecommand{\dimb}   {\mathcal{\MakeUppercase{\nb}}} 
\providecommand{\bN}     {\mathbf{N}}         
\providecommand{\bw}     {\mathbf{w}}         
\providecommand{\bv}     {\mathbf{v}}         
\providecommand{\bq}     {\mathbf{q}}         
\providecommand{\br}     {\mathbf{r}}         
\providecommand{\bJ}     {\mathbf{J}}         
\providecommand{\bqdot}  {\dot{\bq}}          

\providecommand{\bB}     {\mathbf{B}}         

\providecommand  {\bu}  {\mathbf{u}}          
\providecommand  {\nu}  {q}                   
\providecommand  {\dimu}{\mathcal{\MakeUppercase{\nu}}} 
\providecommand  {\bpi} {\boldsymbol{\pi}}    

\providecommand{\bxn}    {\bx_{\nd}}          

\providecommand{\degree}{^\circ}

\providecommand{\ebv}     {\estimated{\bv}}        
\providecommand{\ebw}     {\estimated{\bw}}        
\providecommand{\ebLambda}{\estimated{\bLambda}}   

\providecommand{\bxn}     {\bx_\nd}         
\providecommand{\bun}     {\bu_\nd}         
\providecommand{\bpin}    {\bpi_\nd}       
\providecommand{\ebNn}    {\ebN_\nd}       

\providecommand{\bpir}    {\boldsymbol{\psi}} 

\providecommand{\bL}      {\mathbf{L}}            
\providecommand{\blambda} {\boldsymbol{\lambda}} 

\providecommand{\bsigma}  {\boldsymbol{\sigma}}  
\providecommand{\bC}      {\mathbf{C}}

\renewcommand  {\v    }{\mu}                    
\providecommand{\ev   }{\estimated{\v}}        

\renewcommand  {\r    }{r}                    
\providecommand{\rd}   {\r^*}                 
\providecommand{\brd}  {\br^*}                

\providecommand{\betar}{\beta_r}               
\providecommand{\bxdr} {\bxd_r}                
\providecommand{\bJr}  {\bJ_r}                 
\providecommand{\bAr} {\bA_r}                 

\providecommand{\dimPhi }{\mathcal{\MakeUppercase{p}}} 

\providecommand{\bxd  }{\bx^*}      %

\usepackage[normalem]{ulem}                                        
\usepackage{marginnote}
\usepackage[textwidth=\marginparwidth,colorinlistoftodos]{todonotes}

\colorlet{jb}{red}
\colorlet{mh}{red}
\colorlet{jm}{blue}

\providecommand{\sref}[1]{\S\ref{#1}}
\providecommand{\il}[1]{#1}
\providecommand{\figurename}{Fig.}
\providecommand{\tablename}{Table}

\providecommand{\citeauthor}[1]{\cite{#1}}
\providecommand{\etal}{\textit{et al.}\xspace}

\renewcommand{\baselinestretch}{0.91}
\IEEEoverridecommandlockouts
\begin{document}
\title{\LARGE \bf Exploiting Ergonomic Priors in Human-to-Robot Task Transfer}
\author{Jeevan Manavalan$^{1*}$, Prabhakar Ray \& Matthew Howard
\thanks{%
$^{1}$Jeevan Manavalan, Prabhakar Ray and Matthew J. Howard are with the Centre for Robotics Research, Department of Engineering, King's College London, London, UK. \texttt{jeevan.manavalan@kcl.ac.uk}
}
}

\maketitle

\begin{abstract}
In recent years, there has been a booming shift in the development of versatile, autonomous robots by introducing means to intuitively teach robots task-oriented behaviour by demonstration. In this paper, a method based on \PbD is proposed to learn null space policies from constrained motion data. The main advantage to using this is generalisation of a task by retargeting a systems redundancy as well as the capability to fully replace an entire system with another of varying link number and lengths while still accurately repeating a task subject to the same constraints. The effectiveness of the method has been demonstrated in a 3-link simulation and a real world experiment using a human subject as the demonstrator and is verified through task reproduction on a 7DoF physical robot. In simulation, the method works accurately with even as little as five data points producing errors less than $\mathbf{10^{-14}}$. The approach is shown to outperform the current state-of-the-art approach in a simulated 3DoF robot manipulator control problem where motions are reproduced using learnt constraints. Retargeting of a systems null space component is also demonstrated in a task where controlling how redundancy is resolved allows for obstacle avoidance. Finally, the approach is verified in a real world experiment using demonstrations from a human subject where the learnt task space trajectory is transferred onto a 7DoF physical robot of a different embodiment.

\end{abstract}
\section{Introduction}
\label{s:introduction}
%
\noindent In recent years, there has been a booming shift in the development of versatile, autonomous robots capable of performing increasingly complex tasks. Such robots are expected to enhance the capabilities of ordinary people to introduce automation into their lives by means of intuitively teaching robots task-oriented behaviour by demonstration \cite{surveyRobotArgall,Sena2020}.

\begin{figure}[t!]
      \centering%
      \begin{overpic}[width=.5\linewidth,height=.15\textheight]{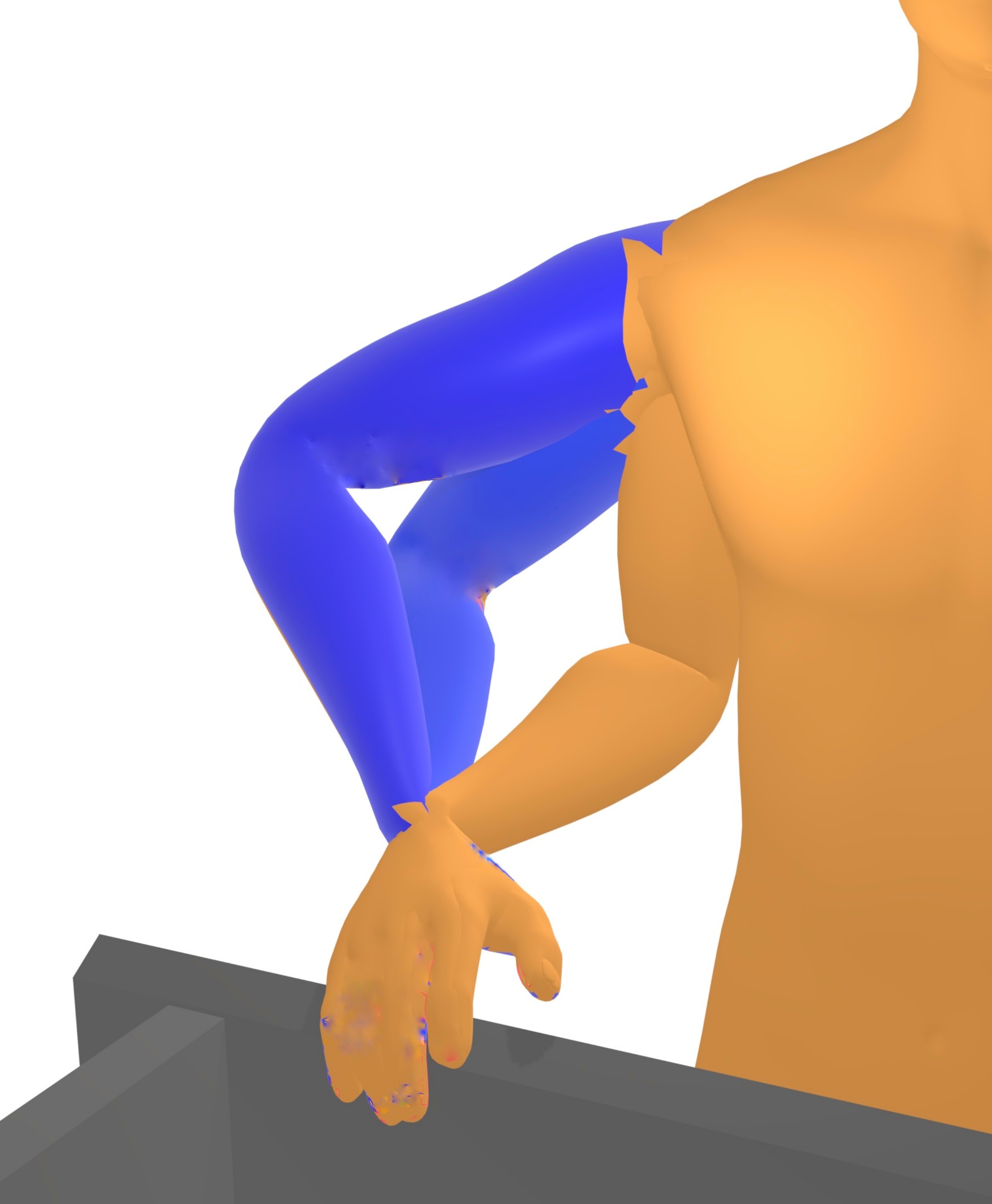}%
      \put(0,75){\ref{f:drawingPosesA}}%
	  \end{overpic}~%
	  \begin{overpic}[width=.5\linewidth,height=.15\textheight]{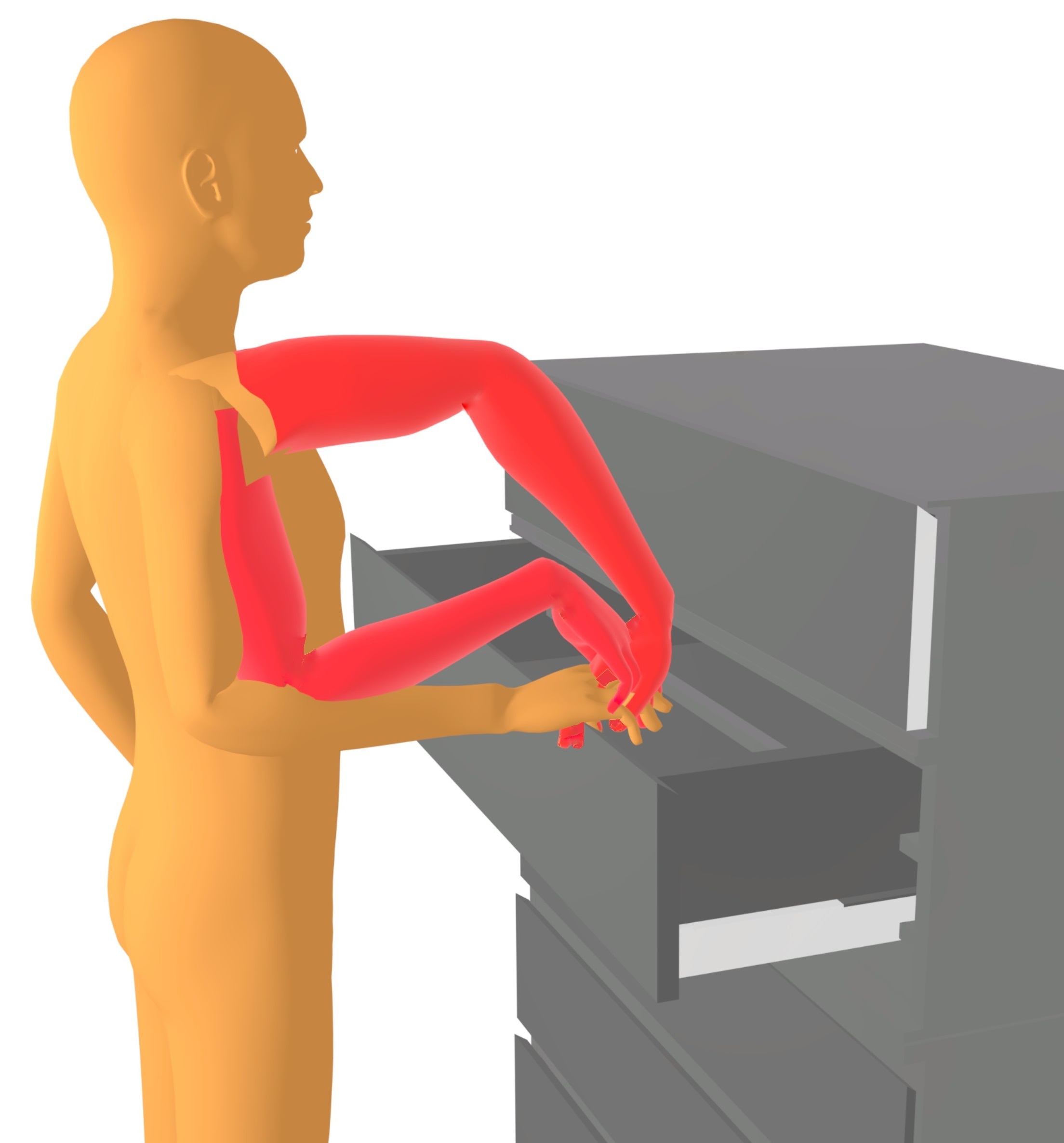}%
  	  \put(70,75){\ref{f:drawingPosesB}}%
	  \end{overpic}%
      \caption{\begin{enumerate*}[label=(\alph*)]%
      		\item\label{f:drawingPosesA} Redundancy in elbow postures when opening a drawer. In absence of other constraints, humans tend to avoid using non-ergonomic postures (shown in blue). %
      		\item\label{f:drawingPosesB} A comfortable pose for a human is different to one maximises manipulability in a robot with different joint limits. %
      	\end{enumerate*} %
      }%
      \label{f:drawingPoses}%
      \vspace{-5mm}%
\end{figure}

When humans perform task-oriented movement, it is often the case that there is a high level of redundancy, with the number of \newabbreviation{\DoF}{DoF} available to execute the task usually much higher than those required \cite{humanReduandancyHolk}. For instance, in the task of opening a drawer (see \fref{drawingPoses}), the primary objective is to \emph{manipulate the drawer from the closed to the open position}, however, there is redundancy in the several possible ways this can be achieved. For example, a human performing this task can adopt different elbow postures, such as flaring it out at various degrees, while still managing to move the drawer (\fref{drawingPoses}\ref{f:drawingPosesA}). 
Having this flexibility is beneficial as it not only allows for multiple ways of achieving the task, but can also enhance efficiency or robustness, thereby enhancing overall performance.

Humans tend to take advantage of this flexibility in predictable, stereotypical ways, commonly, seeking to minimise discomfort or energy expenditure \cite{learningLazyRedundancy}, \cite{movingEffortlesslySoechting}. 
For instance, in human drawer opening, despite the variety of postures that can be taken, it is typical to keep ones wrist straight, to avoid uncomfortable joint flexion, and ones elbow down, to avoid working unnecessarily against gravity. Indeed, such features are codified in human ergonomics literature, to the point that they shape work environments and policy on safe working practice \cite{realtimeergonomicsAli,ergonomicsOfficeTraining,rulaguide2019}. 


Similarly, a multi-\DoF robotic system imitating the human can adopt different joint configurations that are consistent with maintaining the end-effector on the drawer handle, including those that closely match the human's posture. Therefore, the simplest way to have a robot learn is to match it to the human's posture as closely as possible when executing the task. 
%
%
%
However, such an approach neglects the differences in embodiment between human and robot that may lead to sub-optimal performance of the robot \cite{frameworkForTeachingImpedanceBehaviourYuchen,Howard2013a}. For instance, maintaining a posture in which ones wrist joint is kept straight (\fref{drawingPoses}\ref{f:drawingPosesB}), 
while comfortable for the human, may represent a singular posture for the robot that can lead to dangerous unstable movements. Moreover, for a robot with geared, non-backdriveable joints, it  may cost little energy to maintain the elbow in a flared posture, whereas moving the arm to a more human-like, elbow-down posture may actually expend energy unnecessarily. Such cases suggest a more nuanced approach to human-to-robot behaviour transfer is required, that takes explicit account of the stereotypical features of human movement, and the desirability, or otherwise, to reproduce them in a robotic imitator.

To this end, this paper investigates how stereotypical features of demonstrators' posture control can be used to decompose observed behaviour into task-oriented and redundant components of motion. Specifically, it presents a new method for \PbD whereby explicit use of the underlying null space control policy---as determined by the stereotypical or ergonomic features---is used to learn the task and null spaces involved in the behaviour and their underlying constraints \cite{learningspaceLin,towell2010learning}. The latter allows the original behaviour to be retargeted to an imitator robot that has a different kinematic embodiment, by optimising movement according to the robot's own structure without causing any interference with the task goal \cite{learningSingularityAvoidanceJeevan2019}. Numerical and physical evaluations are reported in which the proposed approach is applied to learn constraints in a toy experiment where performance on varying data lengths and noise are evaluated, a simulated 3\DoF experiment where the approach is compared to the state-of-the-art approach, a demonstration on the benefits of retargeting the system to resolve redundancy for obstacle avoidance, task reproduction from a demonstrator system to an imitator of a different embodiment and finally a real world experiment where demonstrations from a human are used to learn and reproduce the task space motions with a Sawyer, a 7\DoF physical robot. The results indicate that learning in this way outperforms several state-of-the-art methods \cite{armesto2017efficientlearning,dependantLearning2016,jeevan2017learningFast,lin2015learning} in its ability to accurately learn the decomposition from relatively little data where in a comparative study the constraint is learnt from one trajectory of length $2\,s$ (100 data points), with minimal assumptions made on the form of the data. 

\section{Background \& Related Work}\label{s:background}


\subsection{Stereotypical Movement} \label{s:StereotypicalDemonstrations}
\noindent In the natural motion of people, it can generally be assumed that the observed movement will be optimised for efficiency according to their embodiment. In the context of humans teaching robots task-oriented movements, this means that postures adopted by a person in the course of a demonstration are likely to not only meet the requirements of the task, but also show stereotypical traits that reflect the demonstrator's embodiment. For instance, human demonstrators will typically adopt postures that avoid working against gravity, or limb flexion or extension away from the resting posture of the limb. The study of postural preferences and stereotypical movement features in humans has long been studied in the field of \emph{human movement science} and, is particularly well documented in the domain of \emph{human ergonomics} \cite{realtimeergonomicsAli,ergonomicsOfficeTraining}.

Note that, these stereotypical features are \emph{secondary to the task}---that is, they will tend to be promoted to seek comfort and minimise fatigue in movement---but are \emph{subject to any applicable task constraints}. This means that they can be inhibited if the task demands it. For example, in drawer opening, the default rest posture of the shoulder is not maintained: since the hand must be lifted to the drawer handle for the task. Furthermore, if maintaining the elbow-down posture during opening \emph{conflicts with the task} (\eg would result in a collision with a obstacle), then task space extends to the elbow elevation and overrides the default behaviour. 

This flexibility is a hallmark of human behaviour that is not currently captured by existing imitation learning approaches and poses an ongoing challenge. Traditional imitation learning approaches tend to treat behaviours as monolithic control policies, and so do not lend themselves well to task-priorised behaviours \cite{imitationLearningSurvey,surveyRobotArgall,computationalSchaal}.

\subsection{Task Prioritised Behaviour}\label{s:taskPioritisedConstraints}
\noindent To better capture this task-prioritised view of behaviour, several studies have recently focused on modelling demonstrations hierarchically, whereby movement is decomposed into the \textit{task space}---the \DoF required for the primary task---and a \textit{null space} (\ie the remaining \DoF). This draws on several well-established hierarchical control schemes, such as Liegeois' redundant kinemetic control scheme \cite{Liegeois1977}, or Khatib's Operational Space Formulation \cite{Khatib1987}.
%
%
%
%

In this view, actions $\bu\in\R^\dimu$ are assumed to take the form
\begin{equation}
\bu(\bx) = \underbrace{\bA^\dagger(\bx)\bb(\bx)}_{\bv}+ \underbrace{\bN(\bx)\bpi(\bx)}_{\bw}
\label{e:problem-solution}
\end{equation}
where $\bx\in\R^\dimx$ represents state (usually represented either in end-effector or joint space) and $\bA(\bx)\in\R^{\dimb\times\dimu}$ is a matrix describing a system of $\dimb$-dimensional constraints
\begin{equation}
\bA(\bx)\bu(\bx)=\bb(\bx)
\label{e:problem-constraint}
\end{equation}
and 
\begin{equation}
\bN(\bx):=\I-\bA(\bx)^\dagger\bA(\bx)\in\R^{\dimu\times\dimu}
\label{e:N}
\end{equation}
is the null space projection matrix that projects the policy $\bpi(\bx)$ onto the null space of $\bA$. Here, $\I\in\R^{\dimu\times\dimu}$ denotes the identity matrix and $\bA^\dagger=\bA^\T(\bA\bA^\T)^{-1}$ is the Moore-Penrose pseudo-inverse of $\bA$. 

In this view, $\bb(\bx)\in\R^{\dimb}$ represents the \emph{task space policy} describing the primary task to be accomplished, and the lumped term $\bv=\bA^\dagger(\bx)\bb(\bx)$ represents that policy projected into the configuration space. $\bpi(\bx)$ represents the \emph{null space policy}, that encapsulates any actions in the configuration space secondary to the task. Note that, it is typically the case that $\bb$ is unknown (since this is the task that should be learnt by demonstration), and $\bA$ (and therefore $\bN$) is also not explicitly known (since this describes the space in which the unknown task is defined). 

The key insight of this paper is that in many cases, \emph{prior knowledge of $\bpi$ may be assumed}, since it commonly represents the \emph{stereotypical features of secondary movements}. Furthermore, as shown in \sref{method}, knowledge of $\bpi$ enables efficient estimation of the other quantities in \eref{problem-solution} ($\bv$ and $\bN$) that can in turn be used to separate out the task-oriented part of the demonstrations, and thereby replace the secondary components with a control policy tailored to the imitator's embodiment.

\subsection{Learning the Decomposition}\label{s:LearningTheDecomposition}
\noindent Several prior studies have examined the possibility of robot learning by demonstration using the representation \eref{problem-solution}-\eref{N}, and in particular the possibility of learning $\bv$ or $\bA$ under the assumption that only $\bx$ and $\bu$ are observable. However, as noted in \sref{StereotypicalDemonstrations}, in many cases such an assumption is \emph{overly stringent} and can result in degraded estimation performance.

Depending on the assumptions made on its representation (see \sref{RepresentationOfA in Spherical Coordinates}), one of several learning methods can be used to estimate $\bA$ \cite{armesto2017efficientlearning,dependantLearning2016,lin2015learning}. However, all of these methods, rely on the ability to separate the lumped task space term $\bv$ (or, equivalently, the null space term $\bw$) from the demonstrations, and learning performance is highly dependant on the quality of the separation. These studies rely on  the same approach, first proposed by Towell \etal \cite{towell2010learning}, that uses variations in the task space policy $\bb$ and consistency in the null space policy $\bpi$ to form an estimate of the separation. This has several limitations in practice.

First, for the separation to work, it can be difficult to ensure that the data is `rich' enough in terms of the variations seen in the task space policy $\bb$. Second, if working with data which contains different several distinct task-spaces, it is important to separate the data into subgroups and learn within each subset individually. However, this diminishes the learning quality as it tends to make less data available within each subgroup. These requirements can hamper the methods' efficacy as increasingly complex systems and constraints are considered. The approach proposed here does not have such prerequisites---instead, it exploits prior knowledge of the control policy $\bpi$, a component that can often be estimated through an understanding of stereotypical behaviour or consideration of human ergonomics. 


\section{Method}\label{s:method}
\noindent In this section, a new method is defined for estimating the null space projection matrix $\bN$ in redundant systems where some prior knowledge of the redundancy resolution strategy is assumed available.

\subsection{Data}\label{s:representationOfData}
\noindent The proposed method works on data given as $\Nd$ pairs of observed states $\bx_n$ and actions $\bu_n$ collected from task-oriented movement demonstrations. It is assumed that \il{\item observations are in the form presented in \eref{problem-solution}, \item $\bA$, $\bN$ and $\bb$ are not explicitly known for any given observation and \item\label{i:knowpi} $\bpi$ is \emph{known} (or a good estimate is available)}. 

\label{s:estimatingPi}
As noted in \sref{background}, assumption \ref{i:knowpi} is reasonable depending on several factors, including the task at hand and the environment. In most circumstances, healthy human demonstrators will tend to perform tasks in a way that \emph{promotes comfort}. In the experiments reported here, this tendency is captured by assuming that the secondary control policy is a point attractor
\begin{equation}
  \bpi(\bx) = \beta(\bxd-\bx)
\label{e:pi-pointattractorMethod} 
\end{equation}
where $\beta$ is a gain matrix and the point of attraction $\bxd$ is a posture that scores highly according to standard ergonomic assessment procedures such as \RULA \cite{rulaguide2019}. While many ergonomic measures are applicable depending on the experiment, RULA is selected as it is quick and simple to classify static postures, focuses on the upper body which is in line with the drawer handling experiments, and provides a constant joint posture range for optimal scoring \cite{rulaMethod}, \cite{RULALynn}.
%

\subsection{Learning the Null Space Projection Matrix}\label{s:learningTheConstraint}
\noindent The proposed method works by exploiting the orthogonality between the task and null space parts in \eref{problem-solution}. Specifically, noting that $\bv^\T\bw=\bw^\T\bv=0$, \eref{problem-solution} can be written
\begin{equation}
\bw^\T \bu= \bw^\T \bv + \bw^\T \bw=\bw^\T \bw
\end{equation}
yielding the identity
\begin{equation}
\bw^\T(\bu-\bw)=0.
\end{equation}
An estimate of $\bN$, and therefore the null space component $\bw$, can be formed by choosing $\ebw=\ebN\bpi$ consistent with this identity by minimising\footnote{For brevity, here, and throughout the paper, the notation $\mathbf{a}_\nd$ is used to denote the quantity $\mathbf{a}$ evaluated on the $\nd$th sample. For example, if $\mathbf{a}$ is a vector quantity computed from the state $\bx$, then $\mathbf{a}_\nd=\mathbf{a}(\bxn)$.}
\begin{equation}
	E[\ebN]=\sum^{\Nd}_{\nd=1} ||\bpin^\T\ebNn(\bun-\bpin)||.
\label{e:constraint-learning}
\end{equation}
\noindent This minimisation problem can be solved using various nonlinear optimisation tools, in this case Matlab's \texttt{fmincon} is used with the interior-point algorithm, which is a nonlinear optimisation solver for constrained multivariable functions.

\subsection{Representation of the Constraints}\label{s:RepresentationOfA in Spherical Coordinates}
\noindent In order to efficiently learn $\ebN$, a suitable representation needs to be selected. The approach chosen here follows that first proposed by Lin \etal \cite{learningspaceLin,lin2015learning} that represents $\ebN$ in terms of an underlying constraint matrix $\ebA$ according to \eref{N}. This has been shown to be effective both for unstructured problems (\ie where the form of the constraint matrix $\bA$ is completely unknown) and for situations where some features (\ie candidate rows) of $\bA$ are available.
\subsubsection{Unit Vector Representation of $\bA$}
Following \cite{lin2015learning}, if the form of $\bA$ is completely unknown, it can be represented using a (potentially, state-dependent) set of $\dimb$ orthonormal vectors
\begin{equation}
	\ebA=\left[\ba_1^\T\,\ba_2^\T\,\cdots\,\ba_\dimb^\T\right]^\T
  \label{eq:ebA}
\end{equation}
where $\ba_\nb= (\a_{\nb,1},\a_{\nb,2},...,\a_{\nb,\dimu})$ corresponds to the $\nb$th constraint in the observations. The latter can be constructed iteratively by selecting vectors orthonormal to one another where the $\nb$th vector has the form
\begin{align}
\a_{\nb,1}       &= \cos\theta_1                                                \nonumber\\
\a_{\nb,2}       &= \sin\theta_1\cos\theta_2                                    \nonumber\\
\a_{\nb,3}       &= \sin\theta_1\sin\theta_2\cos\theta_3                        \nonumber\\
\vdots                                                                          \nonumber\\
\a_{\nb,\dimu-1} &= \prod_{\nu=1}^{\dimu-2} \sin\theta_{\nu}\cos\theta_{\dimu-1}\nonumber\\
\a_{\nb,\dimu}   &= \prod_{\nu=1}^{\dimu-1} \sin\theta_{\nu}.       
\end{align}
The resultant matrix is represented by a total of $\dimtheta=\dimb(2\dimu-\dimb-1)/2$ parameters 
$\btheta=(\theta_1,\theta_2,\cdots,\theta_\dimtheta)^\T$.

\subsubsection{Representation of $\bA$ with Candidate Rows}\label{s:RepresentationOfA for high dimensions}
In the case that prior information about the form constraint matrix is available, this can be incorporated into the estimate using the approach first proposed by \cite{learningspaceLin}. Here, a suitable representation of the constraint matrix is
\begin{equation}
	\ebA = \ebLambda \bPhi
	\label{e:A-matrix}
\end{equation}
where $\ebLambda\in\R^{\dimb\times\dimPhi}$ is a \emph{selection matrix} (to be estimated during learning) and $\bPhi\in\R^{\dimPhi\times\dimu}$ is a (possibly, state-dependent) feature matrix. The rows of the latter can take generic forms such as a series of polynomials, or can contain candidate constraints if there is prior knowledge of those potentially affecting the system. For instance, one may choose $\bPhi=\bJ(\bx)$, the Jacobian of the manipulator, so that $\ebA=\ebLambda\bJ(\bx)$ encodes constraints on the motion of specific degrees of freedom in the end-effector space.

\subsection{Estimating the Components of the Behaviour}\label{s:estVW}
\noindent Once $\ebN$ is estimated, the decomposition of the behaviour into task-oriented and null space parts is straightforward. The null space space component is given as
\begin{equation}
\ebw=\ebN\bpi
\end{equation}
and the task space part is computed as
\begin{equation}
	\ebv=\bu-\ebw.
	\label{e:task-null-separation}
\end{equation}
Note that, given the estimate $\ebA$, $\ebN$ can be computed using \eref{N}, 
and an estimate of the task space policy $\ebb$ can be obtained using \eref{problem-constraint}.
\subsection{Substituting the Non-task oriented Behaviour}\label{s:applicationOfConstraint}
\noindent As noted in \sref{introduction}, in many cases the redundancy resolution strategy seen in demonstrators' task-oriented behaviour may be ill-suited to the robot imitator. The proposed method provides a simple means of \emph{retargeting the behaviour to the robotic system}, while maintaining the task-oriented parts. Specifically, this is achieved by replacing the controls \eref{problem-solution} with
\begin{equation}\label{e:robotU}
\bu = \ebA^\dagger\ebb+\ebN\bpir
\end{equation}
where $\bpir$ is a (possibly, state-dependent) redundancy resolution policy for the robot. For instance, $\bpir$ could be chosen so as to avoid robot-specific joint limits or singularities \cite{learningSingularityAvoidanceJeevan2019}. Alternatively, if the task space trajectory is predictable, $\ev$ can be used in combination with global optimisation in the null space \cite{nakamura1991advanced}.

\section{Evaluation}\label{s:evaluation}
\noindent In this section, the proposed approach is first examined through a toy experiment, then a more complex 3-link planar system, before evaluating its performance in the context of programming by demonstration with a human demonstrator and a 7\DoF physical robot.\footnote{The data supporting this research are openly available from King’s College London at \texttt{http://doi.org/[link to be made available on acceptance]}. Further information about the data and conditions of
access can be found by emailing \texttt{research.data@kcl.ac.uk}.}
\subsection{Toy Problem}\label{s:toyProblem}
\noindent The aim of the first evaluation is to test the robustness of the proposed method using data from a simple, two-dimensional system with a one-dimensional task space. The set up (based on \cite{learningspaceLin}) is as follows.

Constrained motion data is gathered from a two-dimensional system with a one-dimensional constraint $\bA = \boldsymbol{\alpha} \in \mathbb{R}^{1 \times 2}$. Movement in the task space is defined by the constraint matrix and occurs in the direction of the unit vector $\boldsymbol{\hat{\alpha}} = (\cos{\theta},\sin{\theta})$. This direction is selected from a uniform-random distribution $\theta \sim U[0\degree,180\degree]$ at the start of each trial. The task space policy is a linear point attractor  $b(\bx)_i = r^{*} - r, i \in \{1\}$, where $r$ is the position in the task space and  $r^{*}$ is the target point. 
To simulate varying tasks, the task space targets are selected randomly $r^{*} \sim U[-2,2]$ for each trial.

In the below, learning performance is reported for three different secondary control policies $\bpi$, namely,
\begin{enumerate}
\item \textit{A linear policy}: $\bpi(\bx) = -\bL\bar{\bx}$ where $\bar{\bx} \coloneqq (\bx^\T,1)^\T)$ and $\bL=((2,4,0)^\T,(1,3,-1)^\T)^\T)$.
\item \textit{A limit cycle}: $\dot{\rho}=\rho(\rho_0 - \rho^2)$ with radius $\rho_0=0.75\,m$, angular velocity $\phi=1\,rad/s$, where $\rho$ and $\phi$ are the polar representation of the state, \ie $\bx=(\rho \cos \phi,\rho \sin \phi)^\T$.
\item \textit{A non-linear (sinusoidal) policy}:\\$\bpi=(\cos z_1\cos z_2,-\sin z_1\sin z_2)^\T$ where $z_1=\pi x_1$ and $z_2=\pi(x_2 + \frac{1}{2})$.
\end{enumerate}

The training data consists of $150$ data points, drawn uniform randomly across the space $(\bx)_i \sim U[-1,1], i \in \{1,2\}$. For testing, a further 150 data points are used, generated through the same procedure as above. The constraint is learnt by finding a $\btheta$ which minimises \eref{constraint-learning}. In each trial performance is measured using two metrics. First, the \newabbreviation{\NMSE}{NMSE} in the estimated null space component is evaluated\footnote{The notation $\bC=\bA\oslash\bB$ denotes \emph{Hadamard} (element-wise) \emph{division} of $\bA$ by $\bB$, \ie $(\bC)_{ij}=(\bA)_{ij}/(\bB)_{ij}$.}
\begin{equation}\label{e:wNMSE}
E_{\ebw} =
\frac{1}{\Nd}\sum_{\nd=1}^{\Nd}||(\bw_\nd - \ebw_\nd)\oslash\bsigma_\bu||^2.
\end{equation}
where $\bsigma_\bu\in\R^\dimu$ is a vector containing the element-wise standard deviation of the observations $\bu$. Note that, as $\bw_\nd=\bN_\nd\bpi_\nd$ and $\ebw_\nd=\ebN_\nd\bpi_\nd$, this measure is equal to the \NPPE \cite{lin2015learning}.

Second, the normalised error measure \eref{constraint-learning} is evaluated 
\begin{equation}
	E_{\ebN} = \frac{1}{\Nd||\sigma_\bu||^2}\sum^{\Nd}_{\nd=1} ||\bpin^\T\ebNn(\bun-\bpin)||.
\label{e:NNE}
\end{equation}
It indicates the performance of the minimisation function using only known information from given data and is therefore applicable for practical application. 
\footnote{To evaluate the fitness of $\ebv$, noting that $\bv+\bw=\ebv+\ebw$ can be written as $\bv-\ebv=-\bw+\ebw$ returns the identity $\bv-\ebv=-(\bw-\ebw$), where the error in both components are opposites. Thus, results for the fitness of $\ebw$ can also be considered the same for $\ebv$ and therefore $\ebv$ is omitted.} 
The experiment is repeated $50$ times.

\begin{table}[t]\renewcommand{\NPPE}{NPPE\xspace}\renewcommand{\NPOE}{NPOE\xspace}
\caption{Test data \NMSE in $\ebw$ (mean$\pm$s.d.)$\times 10^{-15}$ and $E_{\ebN}$ (mean$\pm$s.d.)$\times 10^{-8}$ over $50$ trials for different $\bpi$. \label{t:resultsTable}}
\centering
\begin{adjustbox}{width=8.5cm}
%

\begin{tabular}{lcc}
\hline
$\bpi$ & $E_{\ebw}$ & $E_{\ebN}$ \\ \hline
Linear & 1.2147 $\pm$ 2.6458 &  0.4263 $\pm$ 1.3396\\
Limit-cycle &  0.5462 $\pm$ 0.7043 & 1.1616 $\pm$ 1.1682\\
Sinusoidal & 0.3020
 $\pm$ 0.5741
 & 1.6685 $\pm$   1.9316\\ \hline
\end{tabular}
%
\end{adjustbox}
\vspace{-0.2cm}
\end{table}

The \NMSE in $\ebw$ and $E_{\ebN}$ are presented in \tref{resultsTable}. 
As can be seen, $\ebN$ is successfully estimated with errors in $\ebw$ less than $10^{-14}$ and $\ebN$ with less than $10^{-7}$ 
for all of the policies considered. These low errors shows that the constraint matrix can be estimated with very high precision if knowledge of $\bpi$ is available. The overall performance of $\ebw$ is seen to be very roughly around twice as accurate as $E_{\ebN}$, which shows that the task and null space components can generally be reproduced with greater accuracy than indicated by just evaluating $E_{\ebN}$.

\begin{figure*}[t!]
      \centering%
      \begin{overpic}[width=1\linewidth]{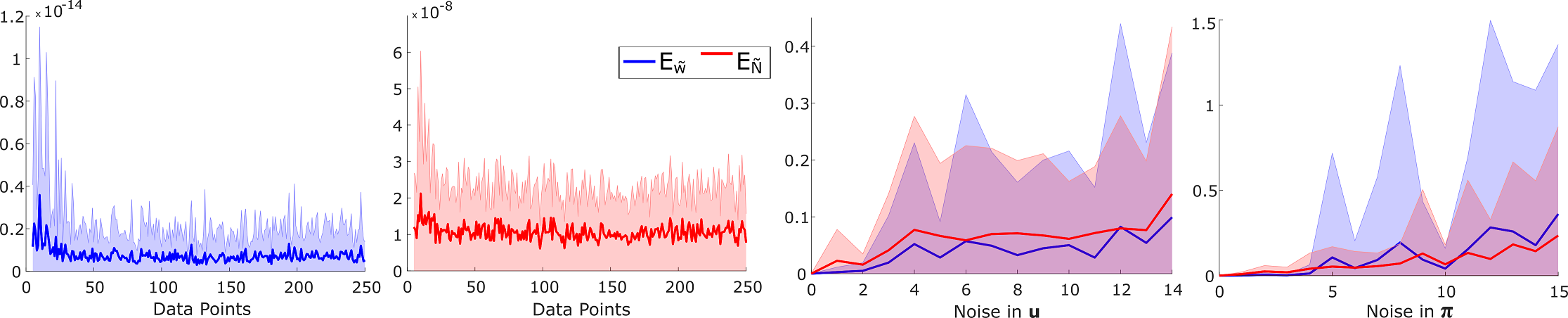}
      \put(0,0){\scriptsize\ref{f:dataToyResults-a}}
      \put(25,0){\scriptsize\ref{f:dataToyResults-b}}
      \put(51,0){\scriptsize\ref{f:dataToyResults-c}}
      \put(76,0){\scriptsize\ref{f:dataToyResults-d}}
      \end{overpic}~\hfill%
      \caption{\NMSE in $\ebw$ and $E_{\ebN}$ (mean$\pm$s.d. over $50$ trials) for %
      	\begin{enumerate*}[label=(\alph*)]%
      		\item\label{f:dataToyResults-a} increasing number of data points for $E_{\ebw}$, 
      		\item\label{f:dataToyResults-b} increasing number of data point for $E_{\ebN}$,
      		\item\label{f:dataToyResults-c} increasing noise levels in $\bu$ and
      		\item\label{f:dataToyResults-d} increasing noise levels in $\bpi$.
      	\end{enumerate*} Mean results are plotted as thick lines and their respective standard deviation are the shaded areas of a similar lighter tone.
      }
      \vspace{-5mm}
      \label{f:dataToyResults}
\end{figure*}
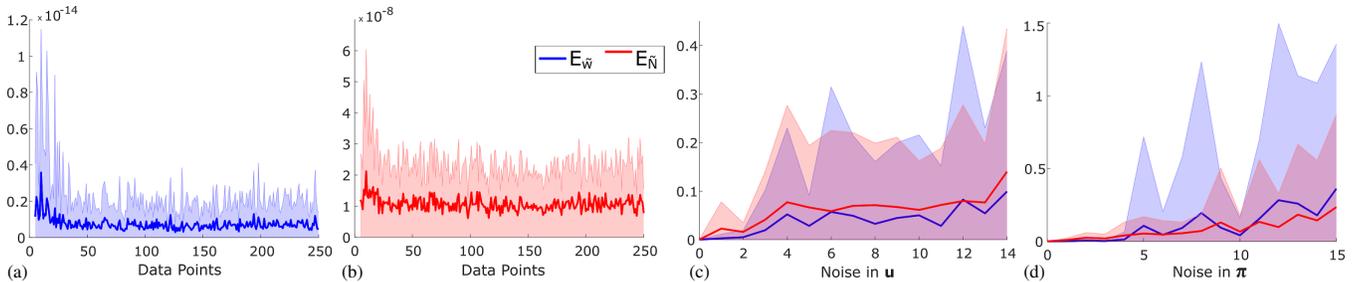

To further characterise the performance of the proposed approach, the experiment is repeated with \il{\item data sets of varying sizes ($5 < \Nd < 250$), \item varying levels of noise in the training data $\bu_\nd$ represented as $N(0,\epsilon \sigma_\bu^2)$ additive white Gaussian noise where $0 < \epsilon < 0.14$ and \item varying levels of noise in the estimated $\bpi_\nd$} where $ 0 < \epsilon < 0.15$ 
for $50$ trials using the limit cycle policy. The latter case simulates error in the assumed $\bpi$ and thereby allows evaluation of the proposed approach in face of an inaccurate estimate of the true underlying redundancy resolution strategy. 
The results are plotted in \fref{dataToyResults}. 

As shown in \fref{dataToyResults}\ref{f:dataToyResults-a}, the \NMSE in $\ebw$ is less than $10^{-14}$ for both mean and standard deviation with as few as five data points. As the number of data points increases, so does the accuracy for minimising $\ebw$ where the performance of the method seems to plateau after around 25 data points. This shows that the approach can learn constraints with as few as five data points and for optimal performance with at least around 25 data points. \fref{dataToyResults}\ref{f:dataToyResults-b} shows errors of less $10^{-7}$ for both mean and standard deviation in $\ebN$. It follows a similar trend to the previous evaluation with respect to accurate learning with as few as five data points and optimal performance after at least around 25 data points. It can also be observed that the learning performance is very roughly half compared to $E_{\ebw}$ which was also observed in \tref{resultsTable}. 
Looking at \fref{dataToyResults}\ref{f:dataToyResults-c}, there is a clear trend with a degrading mean accuracy and greater standard deviation as the noise in $\bu$ increases. The mean error in $\ebw$ stays below 0.1 when $\epsilon <= 0.14$ and for mean error in $\ebN$ when $\epsilon <= 0.13$. It can also been seen that the error in $\ebN$ is greater compared to $\ebw$ in most cases which is in agreement with prior experiments.
Looking at \fref{dataToyResults}\ref{f:dataToyResults-d}, the accuracy decreases, with greater standard deviation, as the error in the assumed $\bpi$ increases. The mean error in $\ebw$ stays below 0.1 when $\epsilon < 0.05$, however when $\epsilon <= 0.1$ only 2 mean values are shown to produce an error above 0.1. The mean $E_{\ebN}$ stays below 0.1 when $\epsilon <= 0.08$ and only has a single instance where the mean value is above 0.1 where $\epsilon <= 0.1$. Comparing $E_{\ebw}$ and $E_{\ebN}$, while both have similar mean performance, the standard deviation of $E_{\ebN}$ is noticeably smaller. This is expected, as $E_{\ebN}$ relies on knowledge of the noisy estimate of $\bpi$ to obtain $\ebN$, whereas $E_{\ebw}$ compares this to the true $\bpi$ to present the error within the estimated null space component. 

\subsection{Simulated Three Link Planar Arm}\label{s:threeLink}
\noindent The aim of the next evaluation is to test the performance of the proposed method on a more complex system with non-linear constraints which simulates a real world system more accurately. The set up is as follows.

Constrained motion data is gathered from a kinematic simulation of a three-link planar robot 
with uniform links of length $10\,cm$. The state and action space refer to the joint angle position and velocities, respectively, \ie $\bx:=\bq \in \mathbb{R}^3$ and $\bu:=\bqdot \in \mathbb{R}^3$. The task space is described by the coordinates $\mathbf{r} = (\r_x,\r_y,\r_\theta)^\T$ referring to the end-effector positions and orientation, respectively. The simulation runs at a rate of $50\,Hz$. 

Joint space motion of the system is recorded as it performs tasks under different constraints in the end-effector space. Specifically, a task constraint at state $\bx$ is described through
\begin{equation}
	\bA(\bx) = \bLambda \bJ(\bx)
	\label{e:AisLJ-matrix}
\end{equation}
where $\bJ \in \mathbb{R}^{3\times 3}$ is the manipulator Jacobian, and $\bLambda\in\mathbb{R}^{3 \times 3}$ is a diagonal selection matrix, with elements $\blambda=(\lambda_x,\lambda_y,\lambda_\theta)^\T$ along the diagonal, indicating which coordinates should be included in ($\lambda_i=1$) or excluded from ($\lambda_i=0$) the task space. In the results reported below, the following six selection matrices are considered:
\il{%
	\item\label{i:lx} $\bLambda_{x}$ where $\blambda=(1,0,0)^\T$,
	\item $\bLambda_{y}$ where $\blambda=(0,1,0)^\T$,
	\item $\bLambda_{\theta}$ where $\blambda=(0,0,1)^\T$,
	\item $\bLambda_{x,y}$ where $\blambda=(1,1,0)^\T$,
	\item $\bLambda_{x,\theta}$ where $\blambda=(1,0,1)^\T$, and
	\item\label{i:lxq} $\bLambda_{x,\theta}$ where $\blambda=(0,1,1)^\T$.
}

To simulate demonstrations of reaching behaviour, the robot end-effector starts from a point chosen uniform-randomly $q_1\sim U[0\degree,10\degree], q_2\sim U[90\degree,100\degree], q_3\sim U[0\degree,10\degree]$ to a task space target $\brd$ following a linear point attractor policy
\begin{equation}
  \bb(\bx) = \brd - \br 
\label{e:b-pointattractor2}
\end{equation}
where $\brd$ is drawn uniformly from $\rd_x \sim U[-1,1]$, $\rd_y\sim U[0,2]$, $\rd_{\theta}\sim U[0\degree,180\degree]$. 
As the secondary control policy a simple point attractor of the form
\begin{equation}
  \bpi(\bx) = \beta(\bxd-\bx)
\label{e:pi-pointattractor3}
\end{equation}
is used, where $\bxd$ is arbitrarily chosen as $x_1=10\degree$, $x_2=-10\degree$, $x_3=10\degree$ and $\beta=1$. 
For each of the cases \ref{i:lx}--\ref{i:lxq} above, $100$ trajectories are generated each containing $50$ data points, with $50$\% of the samples provided for learning and the remainder reserved as unseen testing data. 
Finally, this whole experiment is repeated $50$ times.
\begin{table}[t]
\vspace{4mm}
\captionof{table}{Mean $\pm$s.d. $E_{\ebw}$ and $E_{\ebN}$ on testing data for different null space policies over $50$ trials. The figures for $E_{\ebw}$ are (mean$\pm$s.d.)$\times 10^{-11}$ and for $E_{\ebN}$ are (mean$\pm$s.d.)$\times 10^{-7}$.\label{t:resultsTable3Link}}
\centering
\begin{adjustbox}{width=8cm}
\begin{tabular}{lcc}
\hline
$\bpi$ & $E_{\ebw}$ & $E_{\ebN}$ \\ \hline
$\bLambda_{x}$ & 0.0741 $\pm$ 0.0291 & 1.1527 $\pm$ 2.3654 \\
$\bLambda_{y}$ &  12.1467 $\pm$ 18.6578 & 11.1195 $\pm$ 9.1619 \\
$\bLambda_{\theta}$ & 0.1496 $\pm$ 0.3732 &  0.2445
 $\pm$ 0.1700 \\
$\bLambda_{x,y}$ & 5.6114 $\pm$ 10.3401 & 5.1969 $\pm$ 6.5849 \\
$\bLambda_{x,\theta}$ & 0.0139 $\pm$ 0.0320 & 0.8522 $\pm$ 1.0982 \\
$\bLambda_{y,\theta}$ & 0.0476 $\pm$ 0.1357 & 0.6719 $\pm$ 0.6424 \\ \hline
\end{tabular}
\end{adjustbox}
\vspace{-0.2cm}
\end{table}

The \NMSE in $\ebw$ and $E_{\ebN}$ on the testing data are presented in \tref{resultsTable3Link}. As shown, the constraints are successfully learnt with $E_{\ebw}$ less than $10^{-9}$ and $E_{\ebN}$ less than $10^{-5}$ in all cases. The $E_{\ebw}$ is roughly half of the $E_{\ebN}$ which is in agreement with previous experiments. Overall, the constraint matrix can be accurately estimated using data from the observed demonstrations and knowledge of the control policy, without having to explicitly know how the constraints affect the system's motions.

\begin{figure}[t!]
      \centering%
      \begin{overpic}[width=0.48\textwidth]{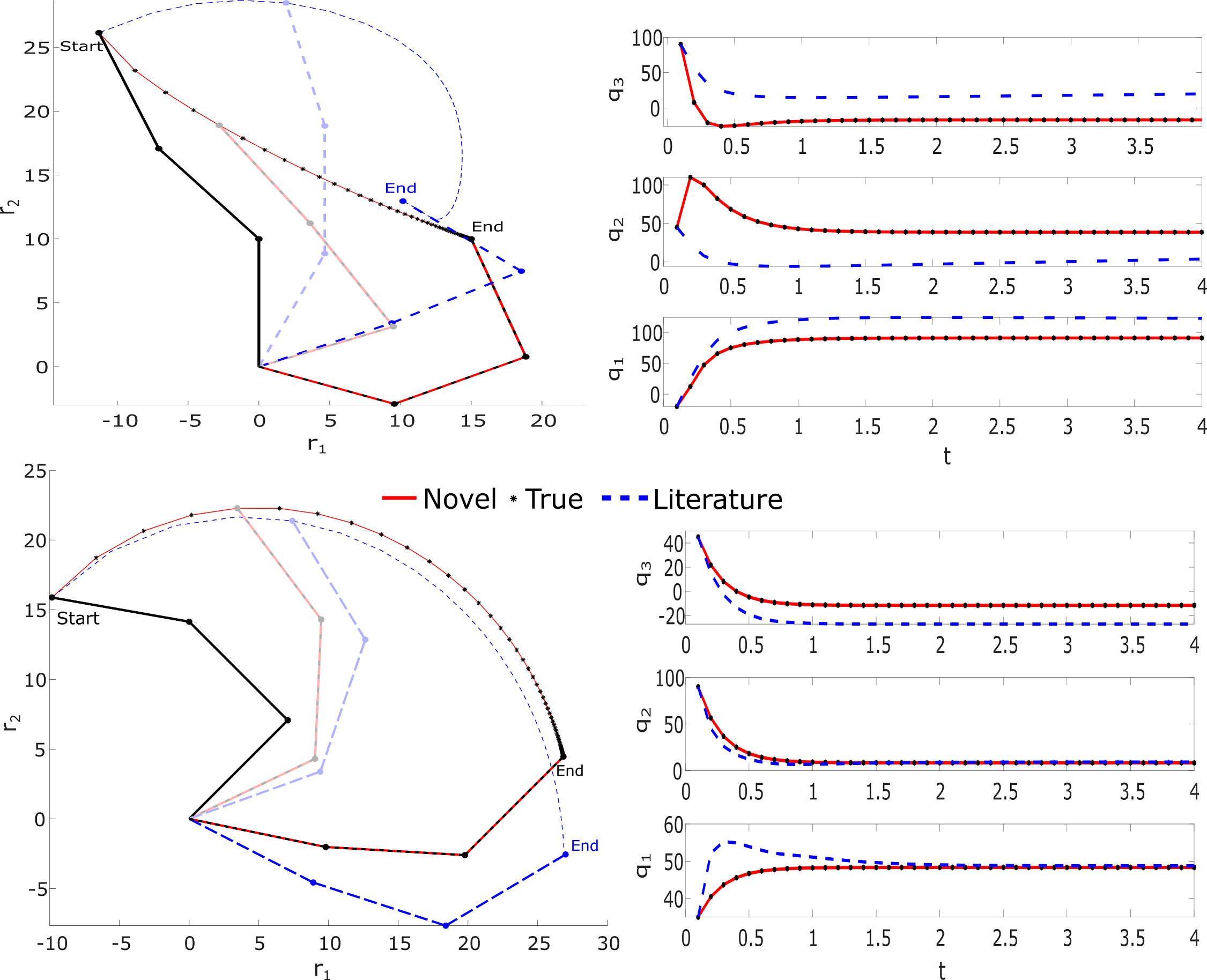}
      \put(0,45){\scriptsize\ref{f:compareLiteratureA}}
      \put(50,45){\scriptsize\ref{f:compareLiteratureB}}
      \put(0,0){\scriptsize\ref{f:compareLiteratureC}}
      \put(52,0){\scriptsize\ref{f:compareLiteratureD}}
      \end{overpic}
      \caption{Reproducing the ground-truth movement (dotted-black) in both learnt task and null space using the proposed method (solid-red) and the state-of-the-art method \cite{towell2010learning} (dashed-blue) to learn the null space component and constraint $\bA$ to obtain $\bb$ \cite{learningspaceLin}. %
      	\begin{enumerate*}[label=(\alph*)]%
      		\item\label{f:compareLiteratureA} Arm visualisation for example task under constraint space $\br = (x, y)$, 
      		\item\label{f:compareLiteratureB} Joint angle positions during example movement in task space $\br = (x, y)$,
      		\item\label{f:compareLiteratureC} Arm visualisation for example task under constraint space $\br = \theta$ and
      		\item\label{f:compareLiteratureD} Joint angle positions during example movement in task space $\br = \theta$.
      	\end{enumerate*} 
      }
      \label{f:compareLiterature}
\end{figure}

To further evaluate the performance of the proposed method, it is compared to the current state-of-the-art. As discussed in \sref{LearningTheDecomposition}, while there are many applicable methods to learn the constraint matrix with acceptable performance including \cite{learningspaceLin}, they all rely on the method proposed in \cite{towell2010learning} for separation of the observed actions. Following \cite{towell2010learning}, \fref{compareLiterature} shows an example of using a learnt constraint to generate a new trajectory. In this experiment, the new trajectory is reproduced using $\ebA$ which is learnt from a separate training data set, $\bu$, $\bx$ and $\bpi$, where the latter three are given. Firstly, training data consists of one trajectory of length $2\,s$ (100 data points) with a random constraint in $x,y$. Using the state-of-the-art approach in \cite{towell2010learning}, $\bu$ is separated into the task and null space components. Now that the null space component is learnt, it is used with $\bu$, $\bx$ and the approach in \cite{learningspaceLin} to obtain $\ebA$. On the other hand, the novel approach uses $\bu$, $\bx$ and $\bpi$ to directly obtain $\ebA$. Now that both approaches have resulted in a learnt constraint, a ground-truth test trajectory is produced subject to the same true constraints in $x,y$ of the training data. Its start pose is $q_1=90\degree, q_2=45\degree, q_3=-20\degree$ and the $x,y$ position of the end-effector moves towards the target point $(15,10)^\T$ which reaches convergence in 4 seconds. To compare the novel and literature approach, both use $\bx$ of this ground-truth data to start at the same position. Both produce $\ebv$ with their respectively learnt $\ebA$ and use this with $\bu$ to estimate $\ebb$ following \eref{problem-constraint}. The literature approach already has $\ebw$ which was obtained using \cite{towell2010learning}. The novel approach uses \eref{N} to obtain $\ebN$ and uses the known $\bpi$ to produce $\ebw$. Both approaches use this information to iteratively reproduce the ground-truth data and similarly run for 4 seconds which is shown in \fref{compareLiterature}\ref{f:compareLiteratureA} and \fref{compareLiterature}\ref{f:compareLiteratureB}. This experiment is repeated for a constraint in $\theta$ shown in \fref{compareLiterature}\ref{f:compareLiteratureC} and \fref{compareLiterature}\ref{f:compareLiteratureD}.


As can be seen in \fref{compareLiterature}\ref{f:compareLiteratureA} and \fref{compareLiterature}\ref{f:compareLiteratureB}, the proposed approach is shown alongside the true policy as well as the current state-of-the-art \cite{towell2010learning} for comparison. The new method follows the true joint trajectories accurately. The state-of-the-art approach on the other hand takes a different route in both the end effector trajectory as well as joint space leading to a different target. 
In \fref{compareLiterature}\ref{f:compareLiteratureC} and \fref{compareLiterature}\ref{f:compareLiteratureD}, the task space target is set as the orientation of the end-effector which moves towards the target angle of $45\degree$. The novel approach accurately reproduces the movements under this 1\DoF constraint unlike the state-of-the-art method in \cite{towell2010learning} and \cite{learningspaceLin}.

\begin{figure}[t!]%
  \centering%
  \begin{overpic}[width=.475\linewidth]{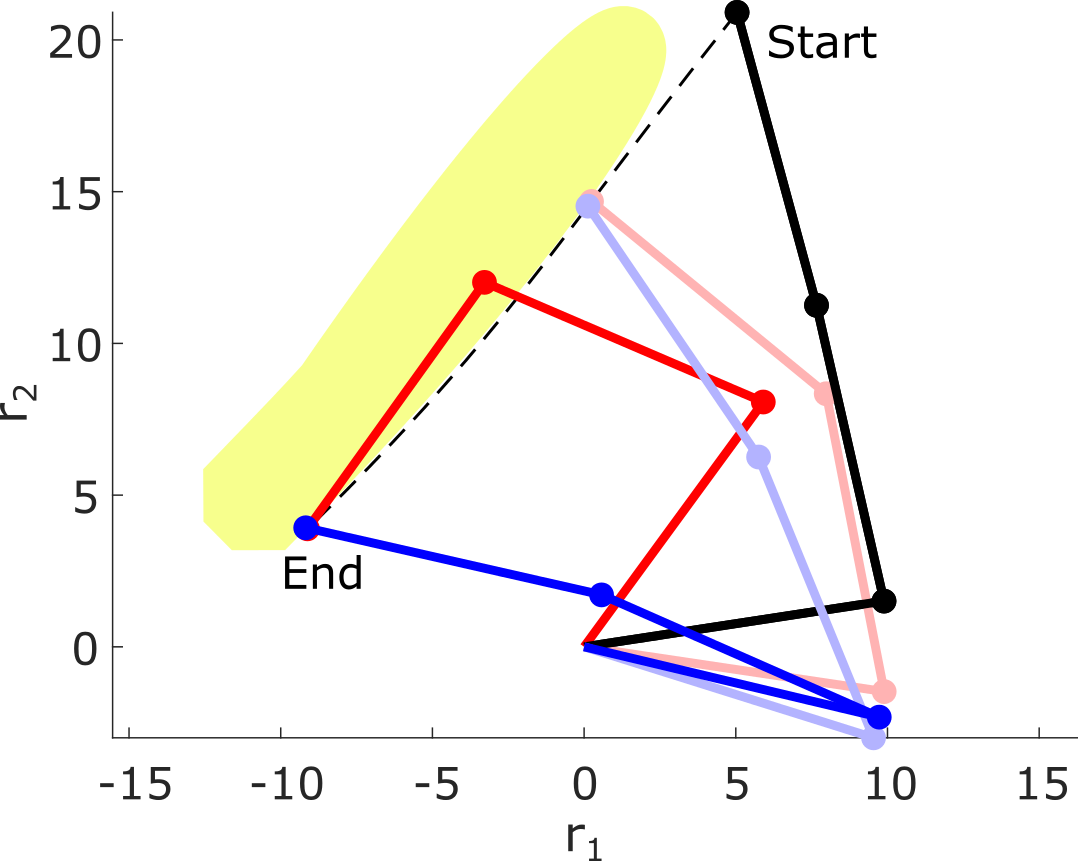}%
  \put(0,0){\scriptsize\ref{f:ThreeToThree}}%
  \end{overpic}~%
  \hfill~%
  \begin{overpic}[width=.475\linewidth]{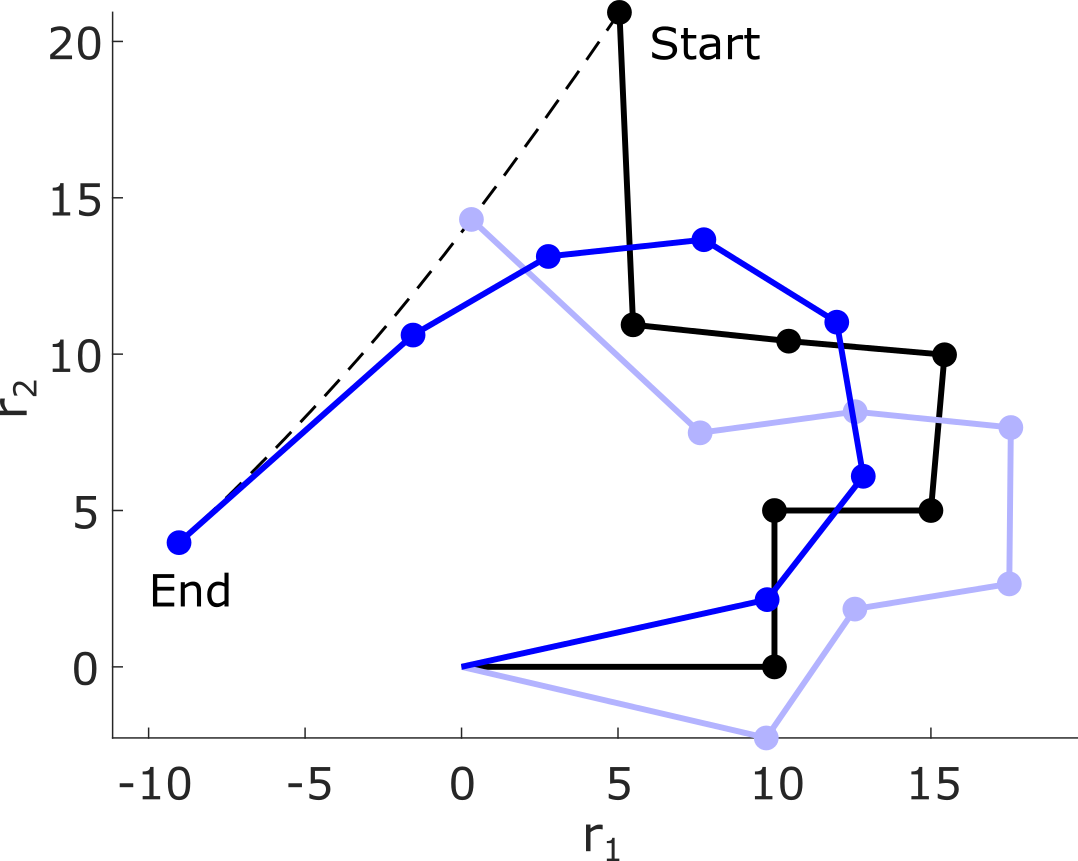}%
      \put(0,0){\scriptsize\ref{f:ThreeToSeven}}%
  \end{overpic}%
  \caption{Retargeting behaviour with an imitator robot %
   	\begin{enumerate*}[label=(\alph*)]%
   		\item\label{f:ThreeToThree} with the same embodiment as the demonstrator but located near to an obstacle, and %
   		\item\label{f:ThreeToSeven} with a different kinematic structure. %
   	\end{enumerate*} %
    The demonstrated movement is illustrated in red, while that of the imitators is in blue. The yellow region indicates an obstacle.%
    }
    \label{f:ThreeToThreeAndSeven}
\end{figure}
 
\begin{figure*}[t!]
      \centering%
 \includegraphics[width=\textwidth, height=4cm]{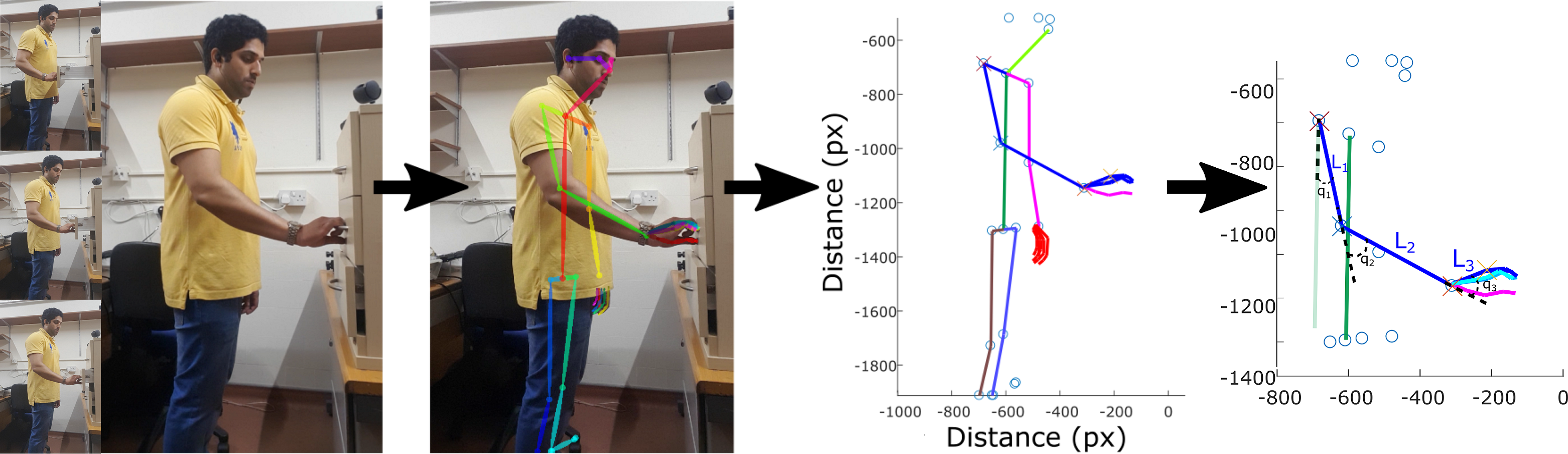}
      \caption{Sample flow of obtaining natural demonstration data from user. Video of the subject is recorded during the action of repeatedly opening and closing drawers of various heights to different lengths. After collection, the video feed is overlaid with a skeleton using Openpose \cite{cao2018openpose} to obtain the positions of body parts such that joint angles and trajectories can be extracted into Matlab. The shoulder, elbow and wrist joint angles are used to learn the constraints contained within the set of demonstrations.}
      \vspace{-5mm}
      \label{f:imageToMatlab}
\end{figure*} 

As mentioned 
in \sref{applicationOfConstraint}, the proposed approach allows \emph{retargeting} of task-oriented behaviour by substituting the demonstrator's redundancy resolution strategy with one better-suited to the robot. 
More concretely, consider the scenario where it is desired to reproduce a demonstrated reaching movement \il{\item with a robot with identical embodiment to the demonstrator, but is located right next to an obstacle (such that there is the risk of collision, see \fref{ThreeToThreeAndSeven}\ref{f:ThreeToThree}), and \item with a robot that has a different kinematic structure to the demonstrator (here, different numbers and lengths of links)}. In the following, the feasibility of retargeting to these scenarios is assessed. As the reaching movement to be reproduced, a typical trajectory is taken from the training data (given in the absence of any obstacles) described above. Specifically, the example chosen uses $\bLambda=\bLambda_{x,y}$, $\bw$ derived from the policy \eref{pi-pointattractor3}, $\brd=(-9.12, 3.89)^\T$ and $\bq=(8.67\degree,94.18\degree,-2.32\degree)^\T$. This movement is \emph{retargeted} by using the learnt $\ebA$ to derive $\ebb$, and then applying \eref{robotU} with a replacement null space control policy.

In the case of the robot located next to an obstacle, retargeting simply consists of selecting an appropriate null-space control policy $\bpir$. Here, $\bpir(\bx) = \betar(\bxdr-\bx)$ is used, where $\betar=5$ and $\bxdr=(-320\degree,100\degree,50\degree)^\T$. 
The resultant movements \emph{with} and \emph{without} retargeting are shown in \fref{ThreeToThreeAndSeven}\ref{f:ThreeToThree} (blue and red figures, respectively). As can be seen, were the arm to directly imitate the demonstration (red figure) a collision would occur (second and third links overlap with the yellow region). This could not only jeopardise the success of the task but also potentially cause damage to the system. In contrast, starting at the same start point, the retargeted controller (blue figure) successfully completes the task (converges to the same target point in end-effector space), however, by resolving its redundancy differently it avoids the collision.

In the case of the robot with the different kinematic structure, retargeting is achieved as follows. As noted above, the constraint in this system is represented in the form \eref{A-matrix} where the feature matrix is selected as the Jacobian of the demonstrator's embodiment (\ie $\bPhi=\bJ$) and the selection matrix $\ebLambda$ is learnt. Since the rows of $\bPhi$ represent meaningful quantities (here, the relationship between the joint space and the end-effector position and orientation) a correspondence is drawn between these and the equivalent quantities for the new arm (\eg if the first row of $\bJ$ relates to the Jacobian for the $\r_x$ coordinate, the corresponding row of the Jacobian $\bJr$ for the imitator is selected), and $\bAr$ is constructed accordingly. Substituting into \eref{robotU} gives the controller
\begin{equation}\label{e:robotU-newJ}
\bu = \bAr^\dagger\ebb+(\I-\pinv{\bAr}\bAr)\bpir.
\end{equation}
In this evaluation, the imitator robot is taken to be a $7$-\DoF arm with link lengths $10$, $5$, $5$, $5$, $5$, $5$ and $10\,cm$, and $\bpir== \betar(\bxdr-\bx)$, where $\betar=1$ and $\bxdr=(-10\degree,-10\degree,-10\degree,-10\degree,-10\degree,-10\degree,-10\degree)^\T$. The start posture is chosen such that the initial end-effector position matches that of the demonstration (\ie $\bq=(0\degree,90\degree,-90\degree,85\degree,90\degree,-1\degree,-81.5\degree)^\T$). The resultant movement is shown in \fref{ThreeToThreeAndSeven}\ref{f:ThreeToSeven}. As can be seen, despite the significant difference in embodiment, the task-oriented part of the movement is effectively reproduced, while the imitator-specific null space controller appropriately handles the added redundancy.

\subsection{Real World Human Arm}
\noindent The aim of this final experiment is to test the performance of the proposed approach in the real world using data from a human demonstrator. The set up is as follows. 

The task chosen for this experiment is to teach a robotic system the skill of opening and closing a three-tiered set of drawers (see \fref{imageToMatlab}). To collect data, the demonstrator stands in front of the drawers at approximately an arm's length distance. The starting state of each drawer is randomised (varying from anywhere between fully closed to completely open). Starting with the top drawer it is moved a random distance in the opening or closing direction. This is repeated for each of the drawers producing a total three trajectories which are used for learning a model. A side view of this is recorded from a single $12$MP phone camera (with a sensor of $1.4\mu$m pixels and aperture of f/$1.7$) placed roughly $1\,m$ away from the demonstrator. 
The video data is then post-processed to overlay a skeleton using Openpose \cite{cao2018openpose} 
to estimate the joint lengths and extract the joint angles and velocities during movement. Constrained motion data of movement in the sagittal plane is gathered from three joints of the demonstrator's arm, by observing flexion and extension of the shoulder, elbow and wrist (as well as abduction and adduction of the wrist depending on the forearms pronation/supination. This yields an average of $11$ frames per trajectory which translates into 11 data points. 

Now that the data is collected, to set it up for learning, the joint angles of the demonstrator are treated as the state $\bx:=\bq \in \mathbb{R}^3$ and the joint velocities as the action $\bu:=\bqdot \in \mathbb{R}^3$. The task space is described by the end-effector coordinates $\br = (\r_x,\r_y,\r_\theta)^\T$ referring to the hand position and orientation, respectively. 
The task constraints are described in the form \eref{AisLJ-matrix}, 
where $\bJ \in \mathbb{R}^{3\times 3}$ is the manipulator Jacobian of the demonstrator. To construct the Jacobian which simulates the demonstrator as a system, the link lengths are calculated from the skeleton in Openpose for each frame. As these can vary from frame to frame depending on obscurities in the demonstrators pose and imprecision of Openpose, the mean of the joint lengths at every frame for the 3 trajectories are used. Moreover, when translating recorded movements from pixels to the $x$ and $y$ axis in Matlab, joint lengths are extremely large where the upper arm measures at around 30 meters. Therefore the scale of these joint lengths are proportionally reduced to around 10cm by dividing each by 300. $\bLambda\in \mathbb{R}^{3 \times 3}$ is the selection matrix specifying the coordinates to be constrained. In this experiment, $\bLambda_{x,y}$ represents the ground truth, since the end-effector (demonstrator's hand) must be maintained at the height $y$ of the drawer handle ($y$ changes in each demonstration depending on the which drawer is being manipulated), and $x$ is the task space in which opening or closing action occurs. It is assumed that the control policy in $\bw$ is resolved by the subject with comfort in mind and moves towards a target joint pose following \eref{pi-pointattractor3} with $\bxd=(-90\degree,90\degree,0\degree)^\T$. This posture is chosen as it lies in the middle of each joints optimal range following RULA, resulting in a high score according to RULA's standard ergonomic assessment procedure \cite{rulaguide2019} (see \sref{background}). 

Since the human demonstrations are modelled as a system with its respective Jacobian matrix, $\bu$, $\bx$ and $\bpi$, it can be evaluated like any other robotic system presented so far \cite{humanArm3DOFModel}, \cite{Humanoidarmkinematicmodeling}.
The experiment is repeated $45$ times yielding a total of $135$ trajectories  ($45$ repetitions for each of the 3 drawers). This is done to verify that the performance of learning the constraints is consistent.

The true decomposition of the behaviour, (\ie $\bv$ and $\bw$) are not known, however they can be estimated using $\bLambda_{x,y}$. 
The initial aspect that can be evaluated is the $E_{\ebw}$, however the variance in $\bu$ is quite small and thus $E[\ebN]$ from \eref{constraint-learning} is reported which is $6.3329\pm5.7832$(mean$\pm$s.d.). Looking at the learnt $\bLambda$, the correct constraints are consistently learnt using the novel approach. To demonstrate this, the learnt constraint is used to produce $\ebb$ and the task-oriented trajectory is reproduced on the Sawyer, a 7\DoF revolute physical robotic system with a maximum reach of 1260mm and precision of $\pm$0.1mm. A closing of a drawer trajectory is selected from the human demonstration data. $\bJ \in \mathbb{R}^{3\times 7}$ is the manipulator Jacobian where a correspondence is drawn between these and the human arms decomposed Jacobian. The start pose is $\bq=(-3.89\degree,42.82\degree,25.48\degree,-76.96\degree,-8.23\degree,32.82\degree,88.93\degree)^\T$. $\bpir== \betar(\bxdr-\bx)$, where $\betar=1$ and $\bxdr=(-70.29\degree, -32.47\degree, -15.92\degree, 53.24\degree, -32.55\degree, -17.48\degree,\newline -68.23\degree)^\T$. The resulting trajectory is presented in \fref{sawyerReproduction}. As shown, the Sawyer is able to reproduce the task space component of closing the drawer using its own embodiment and a different $\bpi$ to resolve redundancy subject to the same task constraints.
\begin{figure}[t]
      \centering
      \includegraphics[width=0.48\textwidth]{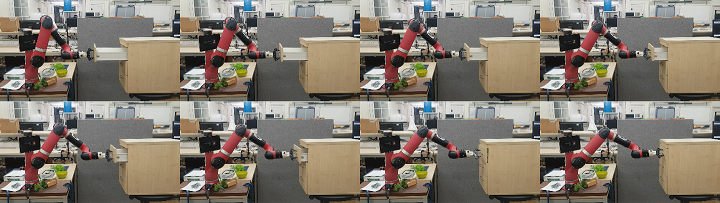}
      \caption{3\DoF Human to 7\DoF Sawyer Robot task transfer}
      \label{f:sawyerReproduction}
      \vspace{-6mm}
\end{figure}

\section{Conclusion}
In this paper, a method based on \PbD is proposed to learn null space policies from constrained motion data. The main advantage to using this is the retargeting of not only the systems redundancy resolution but also the entire system itself with another of a different embodiment which can repeat a task accurately while being subject to the same constraints. On a lesser note, this proposed approach can be used to learn directly from observed actions without the need to decompose motion data into a task and null space component.

The effectiveness of the method has been demonstrated in a simulated toy experiment, a 3-link simulation and a real world experiment using data collected from a human demonstrator which is validated through task-oriented reproduction on a 7\DoF physical robot. All experiments are in agreement that the constraints can be learnt from demonstration. In addition, the evaluations show that the method can \il{\item learn with very little data ($E_{\ebw}$ below $10^{-14}$ with just five data points) and \item handle noise ($E_{\ebw}$ below $10^{-1}$ with normalised additive white gaussian noise below 0.15)}. In a comparative experiment, the approach is shown to outperform the current state-of-the-art approach in a simulated 3\DoF robot manipulator control problem where motions are reproduced using the learnt constraints. It is also used to demonstrate retargeting of a systems null space component to resolve redundancy such that an obstacle can be avoided. Moreover, retargeting through the learnt constraints from the simulated 3\DoF demonstrator to a 7\DoF robot imitator of different embodiment is shown. Finally, the approach is verified in a real world experiment where demonstrations from a human subject are used to consistently learn the constraint matrix, which allows for accurate decomposition of the demonstrated task space and task-oriented reproduction on the Sawyer, a 7\DoF physical robot with a different embodiment.

Future work looks at conducting a study with na\"ive subjects and a more in-depth look at estimating control policies to resolve redundancy.
\bibliographystyle{IEEEtran}
\bibliography{bib/abbreviations_vvshort,bib/bibliography}

\begin{thebibliography}{10}
\providecommand{\url}[1]{#1}
\csname url@samestyle\endcsname
\providecommand{\newblock}{\relax}
\providecommand{\bibinfo}[2]{#2}
\providecommand{\BIBentrySTDinterwordspacing}{\spaceskip=0pt\relax}
\providecommand{\BIBentryALTinterwordstretchfactor}{4}
\providecommand{\BIBentryALTinterwordspacing}{\spaceskip=\fontdimen2\font plus
\BIBentryALTinterwordstretchfactor\fontdimen3\font minus
  \fontdimen4\font\relax}
\providecommand{\BIBforeignlanguage}[2]{{%
\expandafter\ifx\csname l@#1\endcsname\relax
\typeout{** WARNING: IEEEtran.bst: No hyphenation pattern has been}%
\typeout{** loaded for the language `#1'. Using the pattern for}%
\typeout{** the default language instead.}%
\else
\language=\csname l@#1\endcsname
\fi
#2}}
\providecommand{\BIBdecl}{\relax}
\BIBdecl

\bibitem{surveyRobotArgall}
B.~D. Argall, S.~Chernova, M.~Veloso, and B.~Browning, ``A survey of robot
  learning from demonstration,'' \emph{Robotics and Autonomous Systems},
  vol.~57, no.~5, pp. 469--483, 2009.

\bibitem{Sena2020}
A.~Sena and M.H., ``Quantifying teaching behaviour in robot learning from
  demonstration,'' \emph{Int. J. Robotics Res.}, 2020.

\bibitem{humanReduandancyHolk}
H.~Cruse and M.~Brüwer, ``The human arm as a redundant manipulator: The
  control of path and joint angles,'' \emph{Biological cybernetics}, vol.~57,
  pp. 137--44, 02 1987.

\bibitem{learningLazyRedundancy}
R.~Ranganathan, A.~Adewuyi, and F.~A. Mussa-Ivaldi, ``Learning to be lazy:
  Exploiting redundancy in a novel task to minimize movement-related effort,''
  \emph{Journal of Neuroscience}, vol.~33, no.~7, pp. 2754--2760, 2013.

\bibitem{movingEffortlesslySoechting}
J.~Soechting, C.~Buneo, U.~Herrmann, and M.~Flanders, ``Moving effortlessly in
  three dimensions: does donders{\textquoteright} law apply to arm movement?''
  \emph{Journal of Neuroscience}, vol.~15, no.~9, pp. 6271--6280, 1995.

\bibitem{realtimeergonomicsAli}
A.~Shafti, A.~Ataka, B.~U. Lazpita, A.~Shiva, H.~A. Wurdemann, and
  K.~Althoefer, ``Real-time robot-assisted ergonomics,'' in \emph{ICRA}, 2019.

\bibitem{ergonomicsOfficeTraining}
M.~Robertson, B.~Amick, K.~DeRango, T.~Rooney, L.~Bazzani, R.~Harrist, and
  A.~Moore, ``The effects of an office ergonomics training and chair
  intervention on worker knowledge, behavior and musculoskeletal risk,''
  \emph{Applied ergonomics}, vol.~40, pp. 124--35, 04 2008.

\bibitem{rulaguide2019}
\BIBentryALTinterwordspacing
M.~Middlesworth, ``A step-by-step guide to the rula assessment tool,'' 2019.
  [Online]. Available: \url{https://ergo-plus.com/rula-assessment-tool-guide/}
\BIBentrySTDinterwordspacing

\bibitem{frameworkForTeachingImpedanceBehaviourYuchen}
Y.~Zhao, A.~Sena, F.~Wu, and M.~Howard, ``\BIBforeignlanguage{English}{A
  framework for teaching impedance behaviours by combining human and robot
  ‘best practice’},'' in \emph{\BIBforeignlanguage{English}{2018 IEEE/RSJ
  International Conference on Intelligent Robots and Systems (IROS)}}.\hskip
  1em plus 0.5em minus 0.4em\relax IEEE, 10 2018, pp. 3010--3015.

\bibitem{Howard2013a}
M.~Howard, D.~Braun, and S.~Vijayakumar, ``Transferring human impedance
  behavior to heterogeneous variable impedance actuators,'' \emph{{IEEE}
  Transactions on Robotics}, vol.~29, no.~4, pp. 847--862, 2013.

\bibitem{learningspaceLin}
H.~C. Lin, P.~Ray, and M.~Howard, ``Learning task constraints in operational
  space formulation,'' in \emph{ICRA}, 2017, pp. 309--315.

\bibitem{towell2010learning}
C.~Towell, M.~Howard, and S.~Vijayakumar, ``Learning nullspace policies,'' in
  \emph{IROS}, 2010, pp. 241--248.

\bibitem{learningSingularityAvoidanceJeevan2019}
J.~Manavalan and M.~Howard, ``Learning singularity avoidance,'' in
  \emph{IEEE/RSJ International Conference on Intelligent Robots and Systems},
  2019.

\bibitem{armesto2017efficientlearning}
L.~Armesto, J.~Bosga, V.~Ivan, and S.~Vijayakumar, ``Efficient learning of
  constraints and generic null space policies,'' in \emph{ICRA}, 2017, pp.
  1520--1526.

\bibitem{dependantLearning2016}
H.~C. Lin, S.~Rathod, and M.~Howard, ``Learning state dependent constraints,''
  in \emph{IEEE T-Ro}, 2016.

\bibitem{jeevan2017learningFast}
J.~Manavalan and M.~Howard, ``Learning null space projections fast,'' in
  \emph{ESANN}, 2017.

\bibitem{lin2015learning}
H.-C. Lin, M.~Howard, and S.~Vijayakumar, ``Learning null space projections,''
  in \emph{ICRA}, 2015, pp. 2613--2619.

\bibitem{imitationLearningSurvey}
A.~Hussein, M.~Gaber, E.~Elyan, and C.~Jayne, ``Imitation learning: A survey of
  learning methods,'' \emph{ACM Computing Surveys}, vol.~50, 04 2017.

\bibitem{computationalSchaal}
S.~Schaal, A.~Ijspeert, and A.~Billard, ``Computational approaches to motor
  learning by imitation.'' \emph{Philos Trans R Soc Lond B Biol Sci}, vol. 358,
  no. 1431, pp. 537--547, March 2003.

\bibitem{Liegeois1977}
A.~Li\'{e}geois, ``Automatic supervisory control of the configuration and
  behavior of multibody mechanisms,'' \emph{{IEEE} Transactions on Systems,
  Man, and Cybernetics}, vol.~7, pp. 868--871, 1977.

\bibitem{Khatib1987}
O.~Khatib, ``A unified approach for motion and force control of robot
  manipulators: the operational space formulation,'' \emph{IEEE Journal of
  Robotics and Automation}, vol. RA-3, no.~1, pp. 43--53, December 1987.

\bibitem{rulaMethod}
L.~Mcatamney and E.~Nigel~Corlett, ``Rula: A survey method for the
  investigation of work-related upper limb disorders,'' \emph{Applied
  ergonomics}, vol.~24, pp. 91--9, 05 1993.

\bibitem{RULALynn}
L.~McAtamney and E.~N. Corlett, ``Rula: a survey method for the investigation
  of work-related upper limb disorders,'' \emph{Applied Ergonomics}, vol.~24,
  no.~2, pp. 91 -- 99, 1993.

\bibitem{nakamura1991advanced}
Y.~Nakamura, \emph{Advanced Robotics: Redundancy and Optimization}.\hskip 1em
  plus 0.5em minus 0.4em\relax Addison-Wesley Publishing Company, 1991.

\bibitem{cao2018openpose}
Z.~Cao, G.~Hidalgo, T.~Simon, S.-E. Wei, and Y.~Sheikh, ``Open{P}ose: realtime
  multi-person 2{D} pose estimation using {P}art {A}ffinity {F}ields,'' in
  \emph{arXiv preprint arXiv:1812.08008}, 2018.

\bibitem{humanArm3DOFModel}
B.~Lee, ``A mouse with two optical sensors that eliminates coordinate
  disturbance during skilled strokes,'' \emph{Human-Computer Interaction},
  vol.~30, 03 2014.

\bibitem{Humanoidarmkinematicmodeling}
A.~Gams and J.~Lenarcic, ``Humanoid arm kinematic modeling and trajectory
  generation,'' in \emph{IEEE/RAS-EMBS International Conference on Biomedical
  Robotics and Biomechatronics}, 2006, pp. 301 -- 305.

\end{thebibliography}
\end{document}